%% file: paper.tex
\documentclass[11pt,a4paper]{article}
\usepackage[hyperref]{naaclhlt2018}
\usepackage{times}
\usepackage{latexsym}

\usepackage{url}
\usepackage{examples}
\usepackage{multirow}
\usepackage{graphicx}
\usepackage{amsmath}
\usepackage[para,online,flushleft]{threeparttable}
\usepackage{tablefootnote}

\aclfinalcopy

\title{
Enhanced Word Representations for Bridging Anaphora Resolution
}

\author{
Yufang Hou \\
  IBM Research Ireland\\ {\tt yhou@ie.ibm.com}\\
   }

\date{}

\begin{document}
\maketitle

\begin{abstract}
Most current models of word representations (e.g., \emph{GloVe}) have successfully captured fine-grained semantics. However, 
semantic similarity exhibited in these word embeddings is not suitable for resolving bridging anaphora,
which requires the knowledge of associative similarity (i.e., relatedness) instead of semantic similarity information between synonyms or hypernyms. 
We create word embeddings (\emph{embeddings\_PP}) to capture such relatedness
by exploring the syntactic structure of noun phrases. We demonstrate that using \emph{embeddings\_PP} alone achieves around 30\% of accuracy for bridging anaphora resolution
on the ISNotes corpus. Furthermore, we achieve a substantial gain over the state-of-the-art system \cite{houyufang13a} for bridging antecedent selection. 

\end{abstract}

 \input{intro}

\section*{Acknowledgments}
The author thanks the anonymous reviewers for their valuable feedback.

\bibliographystyle{acl_natbib}
\bibliography{../../../bib/lit/lit}

\end{document}

%% file: intro.tex
\section{Introduction}
\label{sec:intro}

Bridging \cite{clarkherberth75,prince81a,gundel93} establishes entity coherence in a text
by linking anaphors and antecedents via various non-identity relations. 
In Example \ref{ex:bridging}, the link between the bridging anaphor (\textbf{the chief cabinet secretary}) and 
the antecedent (\emph{Japan}) establish local (entity) coherence. 

\begin{examples}
\label{ex:bridging}
\item Yet another political scandal is racking \emph{Japan}. On Friday, \textbf{the chief cabinet secretary} announced that \textbf{eight cabinet ministers} had received five million yen from the industry.

\end{examples}

Choosing the right antecedents for bridging anaphors is a subtask of bridging resolution. For this substask, most previous work \cite{poesio04d,lassalle11,houyufang13a}
calculate semantic relatedness between an anaphor and its antecedent based on word co-occurrence count using certain syntactic patterns.   

Most recently, 
word embeddings gain a lot popularity in NLP community because they reflect human intuitions about semantic similarity and relatedness.
Most word representation models explore the distributional hypothesis which states that words occurring in similar contexts have similar meanings \cite{harris54}.
State-of-the-art word representations such as word2vec skip-gram \cite{mikolov13} and GloVe \cite{pennington14} have been shown to perform well across a variety of NLP tasks,
including textual entailment \cite{tim16}, reading comprehension \cite{chendanqi16}, 
and information status classification \cite{houyufang16}.
However, these word embeddings capture both ``genuine'' similarity and relatedness, and they may in some cases be detrimental to downstream performance \cite{Kiela15}.
Bridging anaphora resolution is one of such cases which requires lexical association knowledge instead of semantic similarity information between synonyms or hypernyms. 
In Example \ref{ex:bridging}, among all antecedent candidates, ``\emph{the chief cabinet secretary}'' is the most similar word to the bridging anaphor ``\textbf{eight cabinet ministers}'' but obviously it is not the antecedent for the latter.

In this paper, we explore the syntactic structure of noun phrases (NPs) to derive contexts for nouns in the GloVe model.
We find that the prepositional structure (e.g., \textbf{X} \emph{of} \textbf{Y}) and the possessive structure (e.g., \textbf{Y}\emph{'s} \textbf{X}) are a useful context
source for the representation of nouns in terms of relatedness for bridging relations.

We demonstrate that using our word embeddings based on PP contexts (\emph{embeddings\_PP}) 
alone achieves around 30\% of accuracy on bridging anaphora resolution in the ISNotes corpus, which is 12\% better than the
original GloVe word embeddings. Moreover, adding an additional feature based on \emph{embeddings\_PP} leads to a significant improvement over a state-of-the-art system on bridging anaphora resolution \cite{houyufang13a}.

\section{Related Work}
\label{sec:relatedwork}

\paragraph{Bridging anaphora resolution.} 
Anaphora plays an important role in discourse comprehension.
Different from \emph{identity anaphora} which indicates that a noun phrase refers back to 
the same entity introduced by previous descriptions in the discourse, \emph{bridging anaphora} links anaphors and antecedents via lexico-semantic,
frame or encyclopedic relations.

Bridging resolution has to recognize bridging anaphors and find links to antecedents. There has been a few works tackling full bridging resolution \cite{hahn.coling96,houyufang14}. In recent years, various computational approaches have been developed for 
bridging anaphora recognition \cite{markert12,houyufang13b} and for bridging antecedent selection \cite{poesio04d, houyufang13a}. This work falls into the latter category and we create a new lexical knowledge 
resource for the task of choosing antecedents for bridging anaphors.

Previous work on bridging anaphora resolution \cite{poesio04d,lassalle11,houyufang13a} explore word co-occurence count in certain syntactic preposition patterns to calculate word relatedness.
These patterns encode associative relations between nouns which cover a variety of bridging relations. 
Our PP context model exploits the same principle but is more general. Unlike previous work which only consider a small number of prepositions per anaphor, 
the PP context model considers all prepositions for all nouns in big corpora. It also includes the possessive structure of NPs. 
The resulting word embeddings are a general resource for bridging anaphora resolution. In addition, it enables efficient computation of word association strength through low-dimensional matrix operations.

\paragraph{Enhanced word embeddings.} Recently, a few approaches investigate different ways to improve the vanilla word embeddings.
\newcite{levyomer14} explore the dependency-based contexts in the Skip-Gram model.
The authors replace the linear bag-of-words contexts in the original Skip-Gram model with the syntactic contexts derived from the automatically parsed dependency trees. 
They observe that the dependency-based embeddings exhibit more functional similarity than the original skip-gram embeddings.
\newcite{benjamin17} show that incorporating dependency-based word embeddings into their selectional preference model slightly improve coreference resolution 
performance. 
\newcite{Kiela15} try to learn word embeddings for similarity and relatedness separately by utilizing a thesaurus and a collection of psychological association norms.
The authors report that their relatedness-specialized embeddings perform better on document topic classification than similarity embeddings. 
\newcite{schwartz16} demonstrate that symmetric patterns (e.g, X \emph{or} Y) are the most useful contexts for the representation of verbs and adjectives. 
Our work follows in this vein and we are interested in learning word representations for bridging relations.

\section{Approach}
\label{sec:method}

\subsection{Asymmetric Prepositional and Possessive Structures}
\label{subsec:pp}
The syntactic prepositional and possessive structures of NPs encode a variety of bridging relations between anaphors and their antecedents. 
For instance, \emph{the rear door of that red car} indicates the part-of relation between ``door'' and ``car'', 
and \emph{the company's new appointed chairman} implies the employment relation between ``chairman'' and ``company''.
We therefore extract noun pairs \emph{door--car}, \emph{chairman--company} by using syntactic structure of NPs which
contain prepositions or possessive forms. 

It is worth noting that bridging relations expressed in the above syntactic structures are asymmetric.
So for each noun pair, we keep the head on the left and the noun modifier on the right.
However, a lot of nouns can appear on both positions, such as ``\emph{\textbf{travelers} in the train station}'', ``\emph{\textbf{travelers} from the airport}'', 
``\emph{hotels for \textbf{travelers}}'', ``\emph{the destination for \textbf{travelers}}''. 
To capture the differences between these two positions, we add the postfix ``\_PP'' to the nouns on the left.
Thus we extract the following four pairs from the above NPs: \emph{travelers\_PP--station}, \emph{travelers\_PP--airport}, \emph{hotels\_PP--travelers}, \emph{destination\_PP--travelers}.

\begin{table*}
 \begin{center}
\begin{threeparttable}
\begin{tabular}{l|l|l}
\hline
\textbf{Target Word}& \textbf{\emph{embeddings\_PP}}& \textbf{\emph{GloVe\_Giga}}\\ 
\hline
president &minister, mayor, governor, clinton & vice, presidency, met, former \\ 
          &bush & presidents\\ 
\hline
president\_PP & vice-president\_PP, federation, republic  & \multicolumn{1}{|c}{---}\\
          &usa, corporation & \\ 
\hline
residents &villagers, citizens, inhabitants, families   & locals, villagers, people, citizens\\ 
          &participants& homes \\ 
\hline
residents\_PP & resident\_PP, neighborhood, shemona\tnote{1} & \multicolumn{1}{|c}{---}\\
          &ashraf, suburbs & \\ 
\hline
members &participants, leaders, colleagues, officials& member, representatives, others, leaders\\ 
          &lawmakers& groups\\ 
\hline
members\_PP & member\_PP, representatives\_PP, basij\tnote{2}  & \multicolumn{1}{|c}{---}\\
          &leaders\_PP, community& \\ 
\hline
travelers &travellers, thirsts\_PP, shoppers & travellers, passengers, vacationers\\ 
          &quarantines\_PP, needle-sharing\_PP&tourists, shoppers\\ 
\hline
travelers\_PP & e-tickets, travellers\_PP, cairngorms\tnote{3} & \multicolumn{1}{|c}{---}\\
          &flagstaffs\_PP, haneda\tnote{4}& \\ 
\hline
\end{tabular}
\begin{tablenotes}
\item[1] Shemona is a city in Israel. \item[2] Basij is a paramilitary group in Iran. 
\item[3] Cairngorms is mountain range in Scotland. \item[4] Haneda is an airport in Japan.
\end{tablenotes}\end{threeparttable}
\end{center}
\caption{\label{tab:embeddings} Target words and their top five nearest neighbors in \emph{embeddings\_PP} and \emph{GloVe\_Giga}}
\end{table*}

\subsection{Word Embeddings Based on PP Contexts (\emph{embeddings\_PP})}
\label{subsec:embiddingspp}
Our PP context model is based on GloVe \cite{pennington14}, which obtains state-of-the-art results on various NLP tasks. We extract noun pairs as described in Section \ref{subsec:pp} from the automatically parsed
Gigaword corpus \cite{gigaword5.0data,napoles12}.
We treat each noun pair as a sentence containing only two words and concatenate all 197 million noun pairs in one document.
We employ the GloVe tookit\footnote{\url{https://github.com/stanfordnlp/GloVe}} to train the PP context model on the above extracted noun pairs.
All tokens are converted to lowercase, and words that appear less than 10 times are filtered. This results in a vocabulary of around 276k words and 188k distinct nouns without the postfix ``\_PP''.
We set the context window size as two and keep other parameters the same as in \newcite{pennington14}. 
We report results for 100 dimension embeddings, though similar trends were also observed with 200 and 300 dimensions.

For comparison, we also trained a 100 dimension word embeddings (\emph{GloVe\_Giga}) on the whole Gigaword corpus, using the same parameters reported in \newcite{pennington14}.

Table \ref{tab:embeddings} lists a few target words and their top five nearest neighbors (using cosine similarity) in \emph{embeddings\_PP} and \emph{GloVe\_Giga} respectively.
For the target words ``residents'' and ``members'', both \emph{embeddings\_PP} and \emph{GloVe\_Giga} yield a list of similar words and most of them have the same semantic type as the target word.
For the ``travelers'' example, \emph{GloVe\_Giga} still presents the similar words with the same semantic type, while \emph{embeddings\_PP} generates both similar words and related words (words containing the postfix ``\_PP'').
More importantly, it seems that \emph{embeddings\_PP} can find reasonable semantic roles for nominal predicates (target words containing the postfix ``\_PP''). For instance, ``president\_PP'' is mostly related to
countries or organizations, and ``residents\_PP'' is mostly related to places.


The above examples can be seen as qualitative evaluation for our PP context model. 
We assume that \emph{embeddings\_PP} can be served as a lexical knowledge resource
for bridging antecedent selection.
In the next section, we will demonstrate the effectiveness of \emph{embeddings\_PP} for the task of bridging anaphora resolution.

\section{Quantitative Evaluation}
For the task of bridging anaphora resolution, we use the dataset ISNotes\footnote{http://www.h-its.org/english/research/nlp/download}
  released by \newcite{houyufang13a}.
This dataset contains around 11,000 NPs annotated for information status including 663 bridging NPs
and their antecedents in 50 texts taken from the WSJ portion of the
OntoNotes corpus \cite{ontonotes4.0data}.
It is notable that bridging
anaphors in ISNotes are not limited to definite NPs as in previous work \cite{poesio97b,poesio04d,lassalle11}. 
The semantic relations between anaphor and antecedent 
in the corpus are quite diverse:
only 14\% of anaphors have a part-of/attribute-of relation with the antecedent and only 7\% of anaphors stand in a set relationship to the antecedent. 
79\% of anaphors have ``other'' relation with their antecedents, without
further distinction. This includes 
encyclopedic relations such as \textbf{the waiter} -- \emph{restaurant}  
as well as context-specific relations such as \textbf{the thieves} -- \emph{palms}.

We follow \newcite{houyufang13a}'s experimental setup and reimplement \emph{MLN model II} as our baseline. We first test the effectiveness of \emph{embeddings\_PP} alone
to resolve bridging anaphors. Then we show that incorporating \emph{embeddings\_PP} into \emph{MLN model II} significantly improves the result.

\subsection{Using \emph{embeddings\_PP} Alone}
\label{subsec:eval1}
For each anaphor $a$, we simply construct the list of antecedent candidates $E_a$ using NPs preceding $a$ from the same sentence as well as 
from the previous two sentences. \newcite{houyufang13a} found that globally salient entities are likely to be the antecedents of all anaphors in a text. We approximate this by adding NPs from the first sentence of 
the text 
to $E_a$. This is motivated by the fact that ISNotes is a newswire corpus and globally salient entities are often introduced in the beginning of an article.
On average, each bridging anaphor has 19 antecedent candidates using this simple antecedent candidate selection strategy.

Given an anaphor $a$ and its antecedent candidate list $E_a$, we predict the most related NP among all NPs in $E_a$ as the antecedent for $a$.
The relatedness is measured via cosine similarity between the head of the anaphor (plus the postfix ``\_PP'') and the head of the candidate. 


This simple deterministic approach based on \emph{embeddings\_PP} achieves an accuracy of 30.32\% on the ISNotes corpus. 
Following \newcite{houyufang13a}, accuracy is calculated as the proportion of the correctly resolved bridging anaphors out of all bridging anaphors in the corpus.

We found that using \emph{embeddings\_PP} outperforms using other word embeddings by a large margin (see Table \ref{tab:result1}), including the original GloVe vectors trained on Gigaword and Wikipedia 2014 dump (\emph{GloVe\_GigaWiki14})
and GloVe vectors that we trained on Gigaword only (\emph{GloVe\_Giga}). 
This confirms our observation in Section \ref{subsec:embiddingspp} that \emph{embiddings\_PP} can capture the relatedness between anaphor and antecedent for various bridging relations.

To understand the role of the suffix ``\_PP'' in \emph{embeddings\_PP}, we trained word vectors \emph{embeddings\_wo\_PPSuffix} using the same noun pairs as in \emph{embeddings\_PP}.
For each noun pair, we remove the suffix ``\_PP'' attached to the head noun. We found that using \emph{embeddings\_wo\_PPSuffix} only achieves an accuracy of 22.17\% (see Table \ref{tab:result1}). This indicates 
that the suffix ``\_PP'' is the most significant factor in \emph{embeddings\_PP}.
Note that when calculating cosine similarity based on the first three word embeddings in Table \ref{tab:result1}, we do not add the suffix ``\_PP'' to the head of an bridging anaphor because such words do not
exist in these word vectors.

\begin{table}[t]
\begin{center}
\begin{tabular}{l|l}
\hline
& \textbf{acc}\\ 
\hline
\emph{GloVe\_GigaWiki14}&18.10\\
\emph{GloVe\_Giga}&19.00\\
\emph{embeddings\_wo\_PPSuffix}&22.17\\
\emph{embeddings\_PP}&$\textbf{30.32}$\\
\hline
\end{tabular}
\end{center}
\caption{\label{tab:result1} Results of \emph{embeddings\_PP} alone for bridging anaphora resolution compared to
the baselines. Bold indicates
statistically significant differences over the baselines using randomization test ($p < 0.01$).}
\end{table}

\subsection{\emph{MLN model II} + \emph{embeddings\_PP}}
\begin{table*}[t]
\begin{center}
\begin{tabular}{l|l}
\hline
& \textbf{acc}\\ 
\hline
\emph{MLN model II}&41.32\\
\emph{MLN model II + GloVe\_GigaWiki14}&39.52\\
\emph{MLN model II + embeddings\_wo\_PPSuffix}&40.42\\
\emph{MLN model II + embeddings\_PP}&$\textbf{45.85}$\\
\hline
\end{tabular}
\end{center}
\caption{\label{tab:result2} Results of integrating \emph{embeddings\_PP} into \emph{MLN model II} for bridging anaphora resolution compared to
the baselines. Bold indicates
statistically significant differences over the baselines using randomization test ($p < 0.01$).}
\end{table*}

\emph{MLN model II} is a joint inference framework based on Markov logic networks \cite{domingos09}. In addition to modeling the semantic, syntactic and lexical constraints
between the anaphor and the antecedent (local constraints), it models that:
 \begin{itemize}
   \item semantically or syntactically related anaphors are likely to share the same antecedent (joint inference constraints); 
   \item a globally salient entity is preferred to be the antecedent of all anaphors in a text even if the entity is distant to the anaphors (global salience
   constraints);
   \item several bridging relations are strongly signaled by the semantic classes of the anaphor and the antecedent, e.g., a
 job title anaphor such as \emph{chairman} prefers a GPE or an organization antecedent (semantic class constraints).
 \end{itemize}
Due to the space limit, we omit the details of \emph{MLN model II}, but refer the reader to \newcite{houyufang13a} for a full description.

We add one constraint into \emph{MLN model II} based on \emph{embeddings\_PP}: each bridging anaphor $a$ is linked to its most related antecedent candidate using
cosine similarity. We use the same strategy as in the previous section to construct the list of antecedent candidates for each anaphor.
Unlike the previous section, which only uses the vector of the NP head to calculate relatedness, here we include all common nouns occurring before the NP head as well because they also represent the core
semantic of an NP (e.g., ``\emph{earthquake victims}'' and ``\emph{the state senate}'').

Specifically, given an NP, we first construct a list $N$ which consists of the head and all common nouns appearing before the head, we then 
represent the NP as a vector $v$ using the following formula, where the suffix ``\_PP'' is added to each $n$ if the NP is a bridging anaphor:


\begin{equation} \label{eq:vector}
v = \frac{\sum_{n \in N \;} embeddings\_{PP}_n}{|N|}
\end{equation}

Table \ref{tab:result2} shows that adding the constraint based on \emph{embeddings\_PP} improves the result of \emph{MLN model II} by 4.5\%.
However, adding the constraint based on the vanilla word embeddings (\emph{GloVe\_GigaWiki14}) or the word embeddings without the suffix ``\_PP'' (\emph{embeddings\_wo\_PPSuffix}) slightly decreases the result 
compared to \emph{MLN model II}. 
Although \emph{MLN model II} already explores preposition patterns to calculate relatedness between head nouns of NPs, it seems that the feature based on \emph{embeddings\_PP}
is complementary to the original preposition pattern feature. Furthermore, the vector model allows us to represent the meaning of an NP beyond its head easily.

\section{Conclusions}
We present a PP context model based on GloVe by exploring the asymmetric
prepositional structure (e.g., \textbf{X} \emph{of} \textbf{Y}) and possessive
structure (e.g., \textbf{Y}\emph{'s} \textbf{X}) of NPs. We demonstrate that the resulting word vectors (\emph{embeddings\_PP}) are able to capture
the relatedness between anaphor and antecedent in various bridging relations. In addition, adding the constraint based on \emph{embeddings\_PP} yields a significant improvement over a state-of-the-art system
on bridging anaphora resolution in ISNotes \cite{houyufang13a}.


For the task of bridging anaphora resolution, \newcite{houyufang13a} pointed out that future work needs to explore wider context to resolve context-specific bridging relations.
Here we combine the semantics of pre-nominal modifications and the head by vector average using \emph{embeddings\_PP}. We hope that our embedding resource\footnote{\emph{embeddings\_PP} can be downloaded from \url{https://doi.org/10.5281/zenodo.1211616}.} will facilitate 
further research into improved context modeling for bridging relations.